\documentclass[letterpaper, 10 pt, conference]{ieeeconf}  

\IEEEoverridecommandlockouts                              

\usepackage{amsmath} 
\usepackage{amssymb}  
\usepackage{graphicx}
\usepackage{comment}
\usepackage{algorithm}
\usepackage{algpseudocode}
\usepackage{hyperref}
\usepackage{wrapfig}

\usepackage[dvipsnames]{xcolor}
\usepackage{caption}

\usepackage{tabularx}
\usepackage{multirow}
\usepackage{multicol}
\usepackage{colortbl}

\title{\LARGE \bf
DyPNIPP: Predicting Environment Dynamics for RL-based Robust Informative Path Planning}

\author{Srujan Deolasee$^{1}$, Siva Kailas$^{2}$, Wenhao Luo$^{3}$, Katia Sycara$^{1}$ and Woojun Kim$^{1}$
        \thanks{$^{1}$Srujan Deolasee, Woojun Kim, and Katia Sycara are with the Robotics Institute, Carnegie Mellon University
        {\tt\small \{sdeolase, woojunk, katia\}@cs.cmu.edu}}%
        \thanks{$^{2}$Siva Kailas is with the Department of Computer Science, Georgia Institute of Technology
        {\tt\small \{skailas3\}@gatech.edu}}%
\thanks{$^{3}$Wenhao Luo is with the Department of Computer Science, University of North Carolina at Charlotte
        {\tt\small \{wenhao.luo\}@uncc.edu}}%
}

\begin{document}

\maketitle
\thispagestyle{empty}
\pagestyle{empty}

\begin{abstract}
 Informative path planning (IPP) is an important planning paradigm for various real-world robotic applications such as environment monitoring. IPP involves planning a path that can learn an accurate belief of the quantity of interest, while adhering to planning constraints. Traditional IPP methods typically require high computation time during execution, giving rise to reinforcement learning (RL) based IPP methods. However, the existing RL-based methods do not consider spatio-temporal environments which involve their own challenges due to variations in environment characteristics. In this paper, we propose \texttt{DyPNIPP}, a robust RL-based IPP framework, designed to operate effectively across spatio-temporal environments with varying dynamics. To achieve this, \texttt{DyPNIPP} incorporates domain randomization to train the agent across diverse environments and introduces a dynamics prediction model to capture and adapt the agent actions to specific environment dynamics. Our extensive experiments in a wildfire environment demonstrate that \texttt{DyPNIPP} outperforms existing RL-based IPP algorithms by significantly improving robustness and performing across diverse environment conditions.
\end{abstract}

\section{Introduction}

Informative path planning (IPP) has been actively studied for robotics deployments involving information acquisition, such as autonomous exploration of unknown areas~\cite{liang2023context, cao2024deep}, environment monitoring \cite{hitz2017adaptive}, and target tracking \cite{wang2023spatio}. IPP aims to find a path for autonomous robots that maximizes acquisition of interests (e.g., intensity of fire in fire monitoring) while adhering to resource constraints. Conventional IPP methods typically involve sampling-based path planning using a graph for spatial environments~\cite{hitz2017adaptive, karaman2011sampling, arora2017randomized}. Recently, IPP solvers for spatiotemporal environments, where the interest changes over time, have also been proposed~\cite{jakkala2023multi, kailas2023multi}. Despite their effectiveness, these methods require heavy computation time to determine the path, limiting their applicability for real-world deployment.

\begin{figure}[t]
  \centering
  \includegraphics[width=1\linewidth]
  {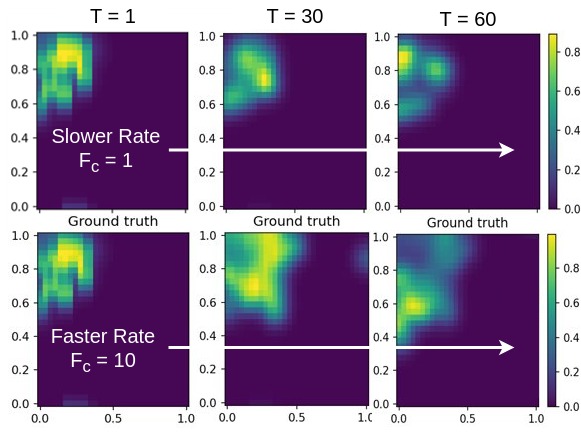}
  \caption{The environment changes in two different dynamics: $F_c=1$ (first row) and $F_c=10$ (second row). $F_c$ is fuel coefficient in the FireCommander simulator \cite{seraj2020firecommander}. Higher $F_c$ causes the fire to spread faster. The higher value (yellow) denotes a higher intensity of fire.}
  \label{fig:envs_veg}
\end{figure}

With the recent success of RL in various domains, RL-based IPP has been actively studied, demonstrating both superior performance and reduced computation time~\cite{catnipp, ruckin2022adaptive, marija_3d}. However, none of the prior works based on RL have considered spatiotemporal environments. Learning in such environments faces an inherent problem of RL: a lack of robustness against variations in environment dynamics \cite{marija_survey}. That is, the agent demonstrates optimal performance only in the environment it was trained on, but not when environment dynamics vary. In addition, even if the agent is trained in an environment with various dynamics, it exhibits suboptimal performance across different environment variations due to over-regularized policies \cite{tiboni2023doraemon}. To understand such problems regarding robustness, let us consider a wildfire domain, where the interest is the fire that spreads over time. Here, the fire growth rate is determined by several characteristics, including fuel. Fig. \ref{fig:envs_veg} shows two environments with different fuel distributions: $F_c=1$ (first row) and $F_c=10$ (second row). These environments have varying environment dynamics due to difference in fuels ($1$: slow, $10$: fast). Consequently, the agent trained in the $F_c=1$ environment has suboptimal performance in the $F_c=10$ environment, and vice versa. The corresponding result in terms of the RMSE error is shown in Table \ref{fig:result_envs_veg}. 
\begin{table}[t]
    \scriptsize
        \centering
        \begin{tabularx}{\columnwidth}{|p{3cm}|>{\centering\arraybackslash}X|>{\centering\arraybackslash}X|}
            \hline
            \multirow{2}{*}{Model} & \multicolumn{2}{c|}{Fuel coefficient} \\
            \cline{2-3}
             & $F_c=1$ & $F_c=10$ \\
             \hline
             CAtNIPP (trained on $F_c$=1) & $13.7 \pm 2.6$ & $25.0 \pm 5.3$ \\
             CAtNIPP (trained on $F_c$=10) & $16.2 \pm 3.6$ & $20.9 \pm 3.0$ \\
             CAtNIPP + DR & $14.9 \pm 2.4$ & $24.0 \pm 5.6$ \\
             \textbf{\texttt{DyPNIPP} (Ours)} & $\textbf{10.3} \pm \textbf{1.6}$ &  $\textbf{14.7} \pm \textbf{2.9}$ \\
             \hline
    \end{tabularx}
    \caption{Performance comparison in terms of RMSE error (lower is better) in the wildfire environment shown in Fig~\ref{fig:envs_veg}.}
    \label{fig:result_envs_veg}
\end{table}
Specifically, CAtNIPP\footnote{This is a prior work on RL-based IPP. This will be discussed later.}~\cite{catnipp} trained on $F_c=10$ performs worse than that trained on $F_c=1$ in the $F_c=1$ environment, implying that the existing RL-based IPP algorithm lacks robustness against variations in environment dynamics.

In this paper, we propose a robust RL-based IPP framework named \texttt{DyPNIPP}, capable of operating effectively across environments with varying dynamics. First, the proposed framework adopts domain randomization (DR), which randomizes environment characteristics (e.g., fuel/vegetation distribution in the wildfire domain) that determine the dynamics, enabling the agent to train across diverse set of environments. However, domain randomization alone is insufficient to train the policy to perform well in diverse environment dynamics. This is because the RL agent is likely trained for averaged dynamics and thus is not capable of inferring specific environment dynamics \cite{tiboni2023doraemon}. Thus, secondly, the proposed method introduces a dynamics prediction model that captures the environment dynamics \textit{implicitly}, allowing the RL policy to recognize the current environment dynamics and adapt the policy accordingly.
In spatiotemporal environments, an IPP agent has to plan a path for an unknown future. A learned inferred dynamics model could greatly improve the capability of the agent to plan informative paths that are robust to unknown variations in the environment. For example, in a wildfire scenario with highly flammable fuel, an agent must spend more time exploring to monitor rapid fire spread, whereas in low-flammability areas, the agent can focus on a more localized area. This adaptability is crucial for an IPP policy as it allows the agent to take actions suited to specific environmental dynamics.
We evaluate the \texttt{DyPNIPP} in the aforementioned wildfire simulation environment with varying fire dynamics. As seen in Table \ref{fig:result_envs_veg}, the results show that domain randomization alone is not sufficient to robustify an RL policy; however, \texttt{DyPNIPP} improves the robustness of the RL policy for the spatio-temporal fire monitoring task.

The main contributions of this work are summarized as
follows: (i) To the best of our knowledge, this is the first work addressing the robustness of RL-based informative path planning in spatio-temporal environments. (ii) We consider key factors for robustness in the wildfire domain, including fuel, vegetation coefficients, and the number of fires. (iii) We demonstrate the effectiveness of our approach across varying conditions---highlighting robustness—and provide analyses of \texttt{DyPNIPP}, along with a real-robot experiment showing the trained model’s practical deployment.

\section{Background and Related Works}

\subsection{Reinforcement Learning}\label{sec:rl}

RL aims to train an agent by interacting with an environment, and the decision making procedure is typically formulated as a Markov decision process (MDP) defined as the tuple $<\mathcal{S}, \mathcal{A}, \mathcal{P}, r, \gamma>$, where 
$\mathcal{S}$ is the state space, $\mathcal{A}$ is the action space, $\mathcal{P}$ is the transition probability (dynamics), $r$ is the reward function, and $\gamma \in [0, 1)$ is the discount factor. At each time step $t$, the agent  generates an action $a_t\in \mathcal{A}$ from a policy $\pi: \mathcal{S}\times \mathcal{A} \rightarrow [0,1]$ based on the given state $s_t\in \mathcal{S}$. The environment then yields the reward $r_t=r(s_t,a_t)$ and the next state $s_{t+1}\sim p(s_{t+1}|s_t,a_t)$. The goal of RL is to learn a policy that maximizes the expected discounted return,  $\mathbb{E}\left[\sum_{t=0}^{\infty}\gamma^t r_t\right]$. One representative algorithm is proximal policy optimization (PPO). PPO consists of an actor and a value function (critic), which estimates expected returns. The agent collects experience, and the value function estimates advantages, indicating how much better or worse an action is compared to the current policy expectation. The actor is then updated to maximize these advantages. PPO uses a clipping method to limit how much the new policy can deviate from the previous one, ensuring stable and efficient learning.

RL has been successfully applied to various robotics tasks, including manipulation~\cite{rl_manipulation}, locomotion~\cite{rl_locomotion}, and path planning~\cite{catnipp} due to its decision-making capabilities. Our work focuses on informative path planning by leveraging the capabilities of RL and addressing its weaknesses in robustness, to achieve a solution that is both effective and robust.

\subsection{Robustness} 
Robustness is considered an essential element for machine learning models to be practical solutions. Robustness can have different definitions depending on the context, such as resistance to adversarial attacks \cite{pattanaik2017robust, zhang2020robust} (i.e., unexpected inputs), and resistance to distribution shifts \cite{chae2022robust, lee2020context}. In this paper, we focus on robustness to general variations in environments within the context of RL. One approach is to introduce a robust MDP, which adds an uncertainty transition set to the MDP, and then use it to estimate the worst-case expected return among perturbed environments \cite{derman2018soft, mankowitz2018learning, mankowitz2019robust}. Another approach is domain randomization \cite{tan2018sim, peng2018sim, slaoui2019robust, zhao2020sim}. Domain randomization typically involves randomizing the environment dynamics to train policies that can adapt to different conditions. For instance, \cite{peng2018sim} leverages domain randomization to develop an RL policy that is robust to variations in environment dynamics, thereby achieving robust sim-to-real transfer. The other approach is modeling environment dynamics via meta-learning \cite{nagabandi2018learning} or supervised learning \cite{lee2020context}. \cite{lee2020context} introduces a separate context-aware dynamics model that infers transitions to adapt to dynamic changes by capturing the environment context. These methods have been applied to several robotics problems, including simulated Mujoco environments and manipulating robotic arms, but have not been considered for the IPP problem.

\subsection{Informative Path Planning}
\label{sec:IPP}
The general IPP problem aims to find an optimal trajectory $\psi^*$ in the space of all available trajectories $\Psi$ to optimize an information-theoretic objective function:
\begin{equation}\label{eq:ipp}
        \psi^* = \operatorname*{arg\,max}_{\psi \in \Psi} I(\psi), s.t.\ C(\psi) \leq B,
\end{equation}
where $I: \psi \rightarrow \mathbb{R^{+}}$ is information gain from the measurements obtained along the trajectory $\psi$, $C: \psi \rightarrow \mathbb{R^{+}}$ maps a trajectory $\psi$ to its associated execution cost, and $B \in \mathbb{R^{+}}$ is the robot's budget limit (e.g., path-length, time, energy). 
In this work, we consider path-length constraints and define the information gain as $I(\psi) = \text{Tr}(P^{-}) - \text{Tr}(P^{+})$, where $\text{Tr}(\cdot)$ denotes the trace of a matrix, and $P^{-}$ and $P^{+}$ represent the prior and posterior covariances, respectively, obtained before and after taking measurements along the trajectory $\psi$, following prior works \cite{catnipp, marija_old, 6224902}. Here, the information gain is computed from sensor measurements taken at fixed intervals along the trajectory $\psi$. Measurements are collected each time the robot travels a fixed distance, and as a result, the total number is determined by the path-length budget $B$.

Optimizing Eq. \ref{eq:ipp}, i.e., solving IPP, is known to be NP-hard. Consequently, computationally tractable approximate methods, such as sampling techniques that explore a complex space by generating random samples of possible solutions, have been proposed~\cite{hitz2017adaptive, karaman2011sampling, arora2017randomized, jones2013receding, yoo2016experimental}. However, these sampling-based methods require heavy computation at test time, which restricts their use in real-world applications.



\textbf{RL-based IPP:} To address the aforementioned problem, RL has been utilized to learn an IPP~\cite{cao2024deep, catnipp,  ruckin2022adaptive, marija_3d, wei2020informative, choi2021adaptive, gadipudi2024offripp}. Most RL-based IPP algorithms are composed of three modules: (1) creating a representation of the entire search map, (2) modeling environmental phenomena, and (3) training an RL agent. One representative example is CAtNIPP~\cite{catnipp}. CAtNIPP trains an RL policy for IPP in static, time-invariant 2D environments.  Before the training starts, CAtNIPP uses a \textit{probabilistic roadmap} (PRM) \cite{prm} that covers the continuous search domain to decrease the complexity of the search space. PRM generates a route graph, $G=(V,E)$, where $V$ and $E$ are sets of nodes and edges, respectively, and each node has $k$ neighbor nodes. Here, the agent is initialized on a randomly chosen node. Next, CAtNIPP leverages GP regression~\cite{seeger2004gaussian} to model the spatial phenomena of the search space, and the output of the GP regression is referred to as the belief of the phenomena. At each time step, the agent observes a measurement at the current node and then uses it to update the belief $\mathcal{GP}(\mu, P)$. Given a set of $n'$ locations $\mathcal{X^*} \subset \mathcal{E}$ at which interest needs to be inferred, a set of $n$ observed locations $\mathcal{X} \subset \mathcal{E}$ and the corresponding measurements set $\mathcal{Y}$, the mean and covariance of the GP is inferred as follows: $\mu = \mu(\mathcal{X^*} + K(\mathcal{X^*, X})[K\mathcal{(X, X)} + \sigma^{2}_{n}I]^{-1}(\mathcal{Y - \mu(X))}$, $P = K(\mathcal{X^*, X^*}) - K(\mathcal{X^*, X})[K(\mathcal{X, X}) + \sigma^{2}_{n}I]^{-1} \times K(\mathcal{X^*, X})^T$, where $K(\cdot)$ is a pre-defined kernel function, $\sigma^{2}_{n}$ is a parameter representing the measurement noise, and $I$ is a $n \times n$ identity matrix. In this paper, we consider Mat\'{e}rn $3/2$ kernel, following the prior works \cite{catnipp, marija_old}.
Lastly, for training the RL policy, the graph augmented by the updated belief—where each node $v_i' = (v_i, \mu(v_i), P(v_i)) \in V$—along with the planning state, which includes the current location, the budget, and the executed trajectory so far, is used as input for the RL policy. Based on such information, the agent chooses one of the neighboring nodes to move to (action). The reward function is based on the previously described information gain, which represents the reduction in uncertainty of GP regression, and is written as $r(t) = (Tr(P^{t-1}) - Tr(P^{t}))/Tr(P^{t-1})$. Summarized above, CAtNIPP trains an RL policy that chooses the neighboring node to move to based on the updated belief-augmented graph to maximize the expected sum of the reduction in uncertainty of GP regression. Additionally, CAtNIPP constructs the RL policy with an attention-based encoder and an LSTM-based \cite{lstm} decoder module.

In addition to CAtNIPP, several RL-based IPP algorithms, including \cite{marija_3d}, where a dynamic graph is proposed to ensure collision-free navigation in 3D environments, have been introduced. The prior works have been shown to be effective in static, time-invariant environments; however, none have considered spatio-temporal environments or variations in environmental dynamics, where the testing environments differ from those on which the agent was trained. To the best of our knowledge, our work is the first to propose an RL-based IPP policy for spatio-temporal environments and rigorously address its robustness.

\section{METHODOLOGY}

\subsection{Motivation}
As described in Sec. \ref{sec:rl}, the standard RL framework assumes a fixed, stationary transition probability, $p(s'|s,a)$, which captures the environment dynamics. Thus, the optimal RL policy in an environment with a specific transition probability may be suboptimal in an environment with a different transition probability. In order to formalize this, we consider the distribution of MDPs, where the transition probability $p(s'|s, a, c)$ is conditioned on environment characteristics $c$. Here, the environment characteristics depend on the domain; for example, in the wildfire domain, $c$ comprises several components that affect fire dynamics, including fuel/vegetation, and wind velocity. 

The existing RL-based IPP algorithms consider the same environment dynamics for both training and testing, using a fixed value for $c$ in both phases~\cite{catnipp, marija_3d}. This, in practice, limits the robustness of the trained RL policy in the face of varying environment dynamics. As we will discuss in Sec. \ref{sec:exp1}, an RL policy trained on specific environment characteristics performs well only in the same environment but poorly in the environments with different characteristics. Since the agent can encounter various environments with different characteristics, it should be capable of handling these variations.



\begin{figure*}
    \centering
    \includegraphics[width=1\linewidth]{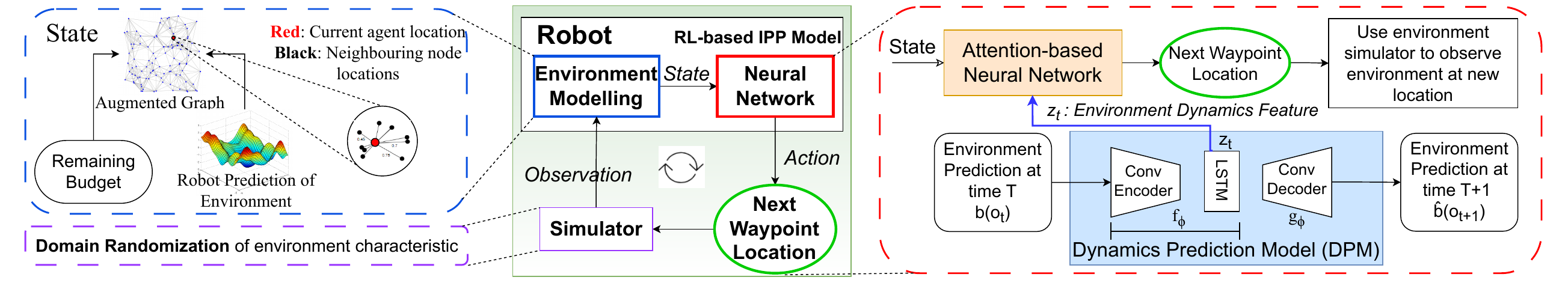}
    \caption{Overview of our approach: (Left) Environment modeling to build the input for the policy and domain randomization for the simulator. (Middle) The overall operation of the proposed IPP algorithm. (Right) Generating the next waypoint (action) using the RL policy and the proposed DPM (blue). DPM predicts the next environment state, and the environment dynamics feature is extracted from the hidden layer to serve as input to the RL policy.}
    \label{fig:network}
\end{figure*}

\subsection{Proposed Method: \texttt{DyPNIPP}}

We aim to train an RL policy for IPP that performs well in environments with varying dynamics---and is robust against the variations. To achieve this, we present \texttt{DyPNIPP}: an environment \textbf{Dy}namcs \textbf{P}rediction \textbf{N}etwork for \textbf{IPP}. The proposed framework comprises of two components: (1) domain randomization (DR), and (2) an additional network along with the RL policy designed to 
capture the environment dynamics by predicting the belief of the next observation. 
Note that \texttt{DyPNIPP} is implemented on top of RL-based IPP algorithms.



\subsubsection{Domain Randomization}

In order to allow the RL agent to encounter a diverse range of environment characteristics, we adopt domain randomization, which involves randomizing the environment dynamics by sampling characteristics from a specified range. This can be interpreted as constructing the environment dynamics by marginalizing the dynamics over the characteristics, written as $p(s'|s,a)=\mathbb{E}_c\left[ p(s'|s,a,c)\right]$. That is, we sample a characteristic from a prior distribution of characteristics (e.g., a uniform distribution $\mbox{Uniform}[1, 10]$ for the fuel/vegetation in the wildfire domain) for each episode and then train the RL policy. As we show in Sec. \ref{sec:exp1}, this domain randomization alone is not enough to find the optimal policy across the characteristics. This is because such a policy is optimal for an environment with marginalized transitions, $\mathbb{E}_c\left[ p(s'|s,a,c)\right]$, implying it is optimal only for the averaged environment, and not for an individual environment with specific characteristics. 

\subsubsection{Dynamics Prediction Model}

To address the aforementioned problem, we introduce a dynamics prediction model (DPM) capable of extracting features of environment dynamics. The proposed DPM uses an encoder consisting of convolutional layers and a LSTM layer, and a decoder involving a dense layer and transposed convolutional layers. The encoder and the decoder parameterized by $\phi_{enc}$ and $\phi_{dec}$ are given by $f(b(o_t), h_t;\phi_{enc})$ and $g(z_t;\phi_{dec})$ respectively. Here, $b(o_t)$ and $h_t$ are the belief of the observation and the hidden vector of the LSTM at time step $t$, respectively. $z_t$ is the output of the encoder referred to as the latent vector. We design the DPM to predict the belief of the next observation 
 based on the belief of the current observation, which is given by GP regression. The corresponding loss function $\mathcal{L}_{DPM}$ is written as:
\begin{align}
    \mathcal{L}_{DPM}(\phi_{enc}, \phi_{dec}) = \mathbb{E} \left[\| y_t - g(z_t;\phi_{dec})\|^2 \right]
\end{align}
where $z_t = f(b(o_t), h_t; \phi_{enc}) ~~ \mbox{and} ~~ y_t = b(o_{t+1}).$ This learning process allows the hidden features of the DPM to represent the environment dynamics. We use the output of the DPM encoder, $z_t$, as an environment-context feature ($z_t$ is 16-dimensional feature).

\subsubsection{RL policy}
Based on the two proposed components, we train an RL policy that selects the next waypoint to move to. Note that DPM can be combined with any RL-based IPP policy. In this work, we use CAtNIPP, which is described in Sec. \ref{sec:IPP}. The key difference is that the environment-context feature of DPM is injected into the RL policy, which is consequently represented as $\pi(a_t|s_t, z_t)$. The detailed architecture of the RL policy to generate action is as follows: The RL policy consists of an attention-based encoder and a attention and LSTM-based \cite{lstm} decoder module. The encoder takes the belief-augmented graph as input to generate a spatially aware embedding for all nodes of the graph. An additional node-embedding is generated using the planning state, which comprises of the remaining budget, executed trajectory, and the augmented graph. We use our environment-context feature $z_t$ as an additional input with these current node-embeddings and pass them to the LSTM block. A decoder is then used to select the most informative neighbor of the robot as an action. The action is selected through a cross-attention network. This attention mechanism computes attention weights by comparing the current node's LSTM features with the neighboring nodes' features. These weights reflect the relevance of each neighbor in the context of the current planning state, and also act as the policy. Since this process uses the budget and the executed trajectory, along with all the spatial and environmental embeddings, the policy is able to suggest actions which are cognizant of the robot's current state, the environment dynamics, and the remaining operational budget. This ensures that the selected actions are informative, spatially-aware, also feasible within the constraints. To train this policy, we use PPO \cite{ppo}, following prior work~\cite{catnipp, marija_3d}.


Summarized above, the overall operation of the proposed method is as follows: (1) we initialize the environment with domain randomization, (2) the robot observes the environmental phenomena and updates the GP, (3) the GP-augmented graph and the remaining budget are fed into the network consisting of the RL policy and DPM, and (4) the environment dynamic feature from the DPM is used by the RL policy. The RL policy generates an action that indicates the next waypoint location for the robot to move to. Repeating this procedure, we train the RL policy and DPM via PPO. This procedure is illustrated in Fig. \ref{fig:network}.





\section{EXPERIMENTS}

\begin{table*}
    \centering
    \scriptsize
    \renewcommand{\arraystretch}{1.2} 
    \begin{tabularx}{\textwidth}{|p{1.8cm}|*{3}{>{\centering\arraybackslash}X} | *{3}{>{\centering\arraybackslash}X} | *{3}{>{\centering\arraybackslash}X}|}
        \hline
        \multirow{3}{*}{Model used} & \multicolumn{9}{c|}{Environment Parameter (Fuel and Vegetation coefficient)} \\
        \cline{2-10}
         & \multicolumn{3}{c|}{budget = 7} & \multicolumn{3}{c|}{budget = 11} & \multicolumn{3}{c|}{budget = 15} \\
         \cline{2-10}
         & $F_c$=1 & $F_c$=5 & $F_c$=10 &  $F_c$=1 & $F_c$=5 & $F_c$=10  &  $F_c$=1 & $F_c$=5 & $F_c$=10  \\
         \hline
          Sampling-based IPP (non-RL) & $523.1 \pm 592.7$ & $551.9 \pm 505.5$ & $550.2 \pm 469.7$ & $949.4 \pm 1161.2$ & $932.0 \pm 870.2$ & $997.4 \pm 916.6$ & $1305.2 \pm 1643.1$ & $1372.7 \pm 1289.7$ & $1367.1 \pm 1142.4$ \\
        CAtNIPP (trained with $F_c=1$) & 460.5 $\pm$ 37.1 & 470.8 $\pm$ 38.3 & 485.6 $\pm$ 38.6 & 450.4 $\pm$ 40.9 & 465.2 $\pm$ 38.6 & 483.7 $\pm$ 41.1 & 442.0 $\pm$ 37.9 & 464.1 $\pm$ 42.7 & 479.3 $\pm$ 45.5 \\ 
        CAtNIPP (trained with $F_c=5$) & 428.2 $\pm$ 28.5 & 427.0 $\pm$ 31.8 & 429.2 $\pm$ 34.4 & 367.2 $\pm$ 27.7 & 372.9 $\pm$ 32.2 & 380.2 $\pm$ 37.4 & 352.8 $\pm$ 26.7 & 367.9 $\pm$ 30.4 & 375.6 $\pm$ 36.4 \\
        CAtNIPP (trained with $F_c=10$) & 496.4 $\pm$ 51.4 & 498.8 $\pm$ 52.2 & 498.6 $\pm$ 50.0 & 464.8 $\pm$ 51.9 & 471.3 $\pm$ 55.5 & 481.7 $\pm$ 53.8 & 457.8 $\pm$ 53.2 & 469.6 $\pm$ 54.9 & 473.5 $\pm$ 50.9 \\ 
        CAtNIPP & 419.3 $\pm$ 30.3 & 422.8 $\pm$ 29.3 & 427.5 $\pm$ 32.6 & 378.7 $\pm$ 25.9 & 387.1 $\pm$ 35.9 & 392.7 $\pm$ 37.0 & 376.4 $\pm$ 28.7 & 389.3 $\pm$ 35.3 & 393.1 $\pm$ 37.0 \\ 
        \textbf{\texttt{DyPNIPP} (Ours)} & \textbf{401.3 $\pm$ 25.5} & \textbf{403.5 $\pm$ 26.4} & \textbf{403.2 $\pm$ 27.3} & \textbf{349.1 $\pm$ 25.3} & \textbf{349.7 $\pm$ 24.9} & \textbf{348.9 $\pm$ 26.5} & \textbf{340.9 $\pm$ 23.0} & \textbf{342.8 $\pm$ 27.1} & \textbf{347.3 $\pm$ 28.7} \\ 
        \hline
    \end{tabularx}
    \caption{Covariance Trace comparison across 200 instances for three budgets; lower values indicate better performance.}
    \label{tab:cov_trace}
\end{table*}

\begin{table*}
    \scriptsize
    \renewcommand{\arraystretch}{1.2} 
    \begin{tabularx}{\textwidth}{|p{1.8cm}|*{3}{>{\centering\arraybackslash}X} | *{3}{>{\centering\arraybackslash}X} | *{3}{>{\centering\arraybackslash}X}|}
        \hline
        \multirow{3}{*}{Model used} & \multicolumn{9}{c|}{Environment Parameter (Fuel and Vegetation coefficient)} \\
        \cline{2-10}
         & \multicolumn{3}{c|}{budget = 7} & \multicolumn{3}{c|}{budget = 11} & \multicolumn{3}{c|}{budget = 15} \\
         \cline{2-10}
         & $F_c$=1 & $F_c$=5 & $F_c$=10 &  $F_c$=1 & $F_c$=5 & $F_c$=10  &  $F_c$=1 & $F_c$=5 & $F_c$=10  \\
         \hline
         Sampling-based IPP (non-RL) & $7.2 \pm 6.8$ & $9.5 \pm 6.5$ & $10.9 \pm 7.9$ & $12.5 \pm 14.1$ & $14.2 \pm 10.9$ & $16.0 \pm 12.8$ & $14.1 \pm 14.8$ & $16.8 \pm 13.6$ & $17.9 \pm 14.2$ \\
         CAtNIPP (trained with $F_c$=1) & $6.2 \pm 1.3$ & $13.4 \pm 3.1$ & $9.5 \pm 2.2$ & $9.5 \pm 2.0$ & $13.2 \pm 2.4$ & $16.1 \pm 4.1$ & $13.7 \pm 2.6$ & $21.8 \pm 4.8$ & $25.0 \pm 5.3$ \\
         CAtNIPP (trained with $F_c$=5) & $6.3 \pm 1.3$ & $8.2 \pm 1.6$ & $0.0 \pm 2.1$ & $9.4 \pm 1.5$ & $12.1 \pm 2.1$ & $13.6 \pm 2.7$ & $15.0 \pm 2.7$ & $18.7 \pm 2.8$ & $18.7 \pm 2.8$ \\
         CAtNIPP (trained with $F_c$=10) & $6.2 \pm 1.5$ & $8.4 \pm 2.1$ & $9.7 \pm 2.6$ & $10.2 \pm 2.7$ & $13.3 \pm 3.4$ & $15.4 \pm 4.1$ & $16.2 \pm 3.6$ & $21.7 \pm 4.7$ & $20.9 \pm 3.0$ \\
         CAtNIPP + DR & $6.1 \pm 1.3$ & $8.2 \pm 1.6$ & $9.9 \pm 2.0$ & $9.1 \pm 1.7$ & $12.9 \pm 2.7$ & $15.3 \pm 3.5$ & $14.9 \pm 2.4$ & $21.4 \pm 4.6$ & $24.0 \pm 5.6$ \\
         \textbf{$\texttt{DyPNIPP}$ (Ours)} & $\textbf{5.2} \pm \textbf{1.0}$ & $\textbf{6.9} \pm \textbf{1.4}$ & $\textbf{7.8} \pm \textbf{1.5}$ & $\textbf{7.7} \pm \textbf{1.3}$ & $\textbf{10.9} \pm \textbf{1.9}$ & $\textbf{11.5} \pm \textbf{2.3}$ & $\textbf{10.3} \pm \textbf{1.6}$ & $\textbf{14.0} \pm \textbf{2.6}$ & $\textbf{14.7} \pm \textbf{2.9}$ \\
         \hline
     \end{tabularx}
    \caption{RMSE comparison across 200 instances for three budgets; lower values indicate better performance.}
    \label{tab:all_cum_rmse}
\end{table*}

\subsection{Environment Setup}

We evaluate the proposed framework 
on a wildfire domain given by the Fire Area Simulator (FARSITE) \cite{farsite} model. We leverage the  FireCommander \cite{seraj2020firecommander} simulator to access the model for generating the spatio-temporal environment. The fire's growth rate (i.e. fire propagation velocity) is a function of fuel and vegetation coefficient ($F_c$), wind speed ($U_c$), and wind azimuth ($\theta_c$). The first-order firespot dynamics $\dot{q_t}$ are estimated for each propagating spot $q_t$, where $q$ is a 2-dimensional vector representing the X-Y coordinates, by $\dot{q_t} = C(F_c, U_c)\mathcal{D}(\theta_c)$, where $\mathcal{D}(\theta_c)$ = [sin($\theta_c$), cos($\theta_c$)]. 
Here, $C(F_c, U_c)$ can be calculated as: $C(F_c, U_c) = F_c \left( 1 - LB(U_c)/(LB(U_c) + \sqrt{GB(U_c)}) \right)$,
where $LB(U_c) = 0.936e^{0.256U_c} + 0.461e^{-0.154U_c} - 0.391$ and $GB(U_c) = LB(U_c)^2 -1$. Note that the fuel and vegetation coefficient $F_c$ is the most dominant factor affecting the fire spread. 
To evaluate the algorithms in terms of robustness, we consider environments with varying the fuel/vegetation coefficient $F_c$ and the number of fire origins within the environment.



\subsection{Training Details}
In the FireCommander simulator, both higher $F_c$ and $U_c$ result in a higher fire spread rate ($F_c, U_c > 0$). Since a higher $F_c$ implies highly flammable fuel, it also means that the fuel can get exhausted more quickly, resulting in a quicker decay of fire. Although theoretically both coefficients can be any positive numbers, 10 is given as the recommended limit for both. For all our experiments, we fix the wind speed $U_c$ (weaker factor) at 5 and vary the fuel and vegetation coefficient $F_c$ (dominant factor) to train and test for robustness. The wind direction is chosen randomly for each episode. 
We use domain randomization to make our policy robust to changes in these parameters which impact the fire spread dynamics. While training our policy, we randomly choose a $F_c$ in between $[1, 10]$ for every episode (unless otherwise stated). Based on these sampled environment parameters, we use FireCommander to generate the spatio-temporal wildfire environment and normalize the field dimensions to construct the true interest map of a unit square $[0,1]^2$ size. Since there are no initial observations of the environment, the robot's belief starts as a uniform distribution $\mathcal{GP}(0, P^0), P^{0}_{i,i} = 1.$ The start and destination positions are randomly generated in $[0,1]^2$. We train our policy with a fixed number (200) of nodes for our graph, where the number of neighboring nodes is fixed to $k = 20$, and the budget is randomized between $[7, 9]$. Note that the graph is reinitialized for each episode. 
A measurement is obtained every time the robot has traveled $0.2$ units from the previous measurement. We set the max episode length to $256$ time steps, and the batch size to $32$. We use the Adam optimizer with learning rate $10^{-4}$, which decays every $32$ steps by a factor of $0.96$. 

\texttt{DyPNIPP} trains a RL agent on top of CAtNIPP~\cite{catnipp} utilizing PPO \cite{ppo} for the training. For each training episode, PPO runs $8$ iterations. Our model is trained on a workstation using AMD EPYC 7713 CPU, and a single NVIDIA RTX 6000 Ada Gen GPU. We use Ray \cite{moritz2018ray} to distribute the training process and run 16 IPP instances in parallel, thus requiring just around 2 hours to converge.

\begin{table*}[t]
    \centering
    \scriptsize
    \renewcommand{\arraystretch}{1.2} 
    \begin{tabularx}{\textwidth}{|p{2cm}|*{3}{>{\centering\arraybackslash}X} | *{3}{>{\centering\arraybackslash}X} | *{3}{>{\centering\arraybackslash}X}|}
        \hline
        \multirow{3}{*}{Model used} & \multicolumn{9}{c|}{Environment Parameter (Fuel and Vegetation coefficient)} \\
        \cline{2-10}
         & \multicolumn{3}{c|}{n(fires) = 1} & \multicolumn{3}{c|}{n(fires) = 3} & \multicolumn{3}{c|}{n(fires) = 5} \\
         \cline{2-10}
         &  $F_c$=1 & $F_c$=5 & $F_c$=10  &  $F_c$=1 & $F_c$=5 & $F_c$=10  &  $F_c$=1 & $F_c$=5 & $F_c$=10  \\
         \hline
         CAtNIPP + DR & $10.3 \pm 2.3$ & $14.5 \pm 3.0$ & $16.5 \pm 3.4$ & $13.9 \pm 2.6$ & $16.5 \pm 3.6$ & $19.7 \pm 2.9$ & $14.0 \pm 2.6$ & $18.3 \pm 3.1$ & $20.5 \pm 3.7$\\
         \textbf{\texttt{DyPNIPP} (Ours)} & $\textbf{9.3} \pm \textbf{2.0}$ & $\textbf{11.9} \pm \textbf{3.1}$ & $\textbf{13.0} \pm \textbf{3.3}$  & $\textbf{12.2} \pm \textbf{2.6}$ & $\textbf{14.4} \pm \textbf{3.6}$ & $\textbf{15.0} \pm \textbf{3.8}$ & $\textbf{13.4} \pm \textbf{2.7}$ & $\textbf{15.4} \pm \textbf{3.4}$ & $\textbf{15.5} \pm \textbf{3.2}$ \\
         \hline
    \end{tabularx}
    \caption{RMSE comparison  with respect to the number of fires; policy trained on environment with up to 3 fires.}
    \label{tab:num_fire}
\end{table*}

\begin{table*}[t]

    \centering
    \scriptsize
    \renewcommand{\arraystretch}{1.2} 
    \begin{tabularx}{\textwidth}{|p{2cm}|*{3}{>{\centering\arraybackslash}X} | *{3}{>{\centering\arraybackslash}X} | *{3}{>{\centering\arraybackslash}X}|}
        \hline
        \multirow{3}{*}{Prediction} & \multicolumn{9}{c|}{Environment Parameter (Fuel and Vegetation coefficient)} \\
        \cline{2-10}
         & \multicolumn{3}{c|}{budget = 7} & \multicolumn{3}{c|}{budget = 11} & \multicolumn{3}{c|}{budget = 15} \\
         \cline{2-10}
         &  $F_c$=1 & $F_c$=5 & $F_c$=10  &  $F_c$=1 & $F_c$=5 & $F_c$=10  &  $F_c$=1 & $F_c$=5 & $F_c$=10  \\
         \hline
         $b(o_t)$ & $7.2 \pm 1.2$ & $8.9 \pm 1.5$ & $9.9 \pm 1.5$ & $10.8 \pm 1.6$ & $13.7 \pm 1.9$ & $14.4 \pm 2.1$ &  $11.3 \pm 2.1$ & $14.8 \pm 2.8$ & $15.4 \pm 3.2$ \\
         $b(o_{t+1}) - b(o_t)$ & $5.3 \pm 1.1$ & $6.8 \pm 1.4$ & $7.9 \pm 1.5$ & $8.0 \pm 1.4$ & $10.7 \pm 1.9$ & $11.7 \pm 2.3$ & $10.5 \pm 1.9$ & $14.2 \pm 3.0$ & $14.7 \pm 3.0$ \\
         $b(o_{t+1})$ (\textbf{Ours}) & $5.2 \pm 1.0$ & $7.0 \pm 1.4$ & $7.8 \pm 1.5$ & $7.8 \pm 1.3$ & $10.9 \pm 1.9$ & $11.5 \pm 2.3$ & $10.3 \pm 1.6$ & $14.0 \pm 2.6$ & $14.7 \pm 2.9$ \\
         \hline
          
    \end{tabularx}
    \caption{Ablation study on the design of our dynamics prediction model}
    \label{tab:pred_ab}
\end{table*}

\subsection{Experimental Results}

In this subsection, we examine (1) whether \texttt{DyPNIPP} enhances the robustness against environment variations, including the fuel coefficient $F_c$ and the number of fires, (2) the effectiveness of our DPM design, and (3) how the environment-context latent feature varies with respect to the dynamics.

\subsubsection{Performance Comparison of Variation in Fuel Coefficient}\label{sec:exp1}


We include three main baselines: First, we include three CAtNIPP policies, each with a spatio-temporal GP, trained on a fixed $F_c \in \{1, 5, 10\}$. Second, we include CAtNIPP combined with domain randomization, where $F_c$ is randomly sampled between $[0, 10]$ for every episode. We refer to this method as CAtNIPP + DR in Table \ref{tab:all_cum_rmse}. Lastly, we include a sampling-based informative planner, where the target location is selected using the following objective function: highest predictive entropy minus the distance between the target location and the robot's location. For this method, we provide 100 observations at randomly sampled locations as initial data to the spatio-temporal GP. 
\textbf{Performance Metrics: }
We compare \texttt{DyPNIPP} with these baselines across three environments, each characterized by a different $F_c \in {1, 5, 10}$, and three different budgets ($B \in {7, 11, 15}$). We use two metrics for evaluation: the covariance trace $Tr(P)$ and the cumulative root mean squared error (RMSE). The covariance trace provides information about the confidence of the robot's GP model—representing the uncertainty remaining at the end of the mission, which aligns with the IPP objective described in \sectionautorefname~\ref{sec:IPP}. Since this does not necessarily correlate with the accuracy of the environment prediction, we introduce the RMSE between the environment prediction and the actual ground truth after every step of the agent's trajectory. We provide the results in terms of the metrics in Table ~\ref{tab:cov_trace} and \ref{tab:all_cum_rmse}. Based on the results, we observe the following:


$\bullet$ \textbf{The prior RL-based IPP algorithm lacks robustness}: CAtNIPP trained on a specific $F_c$ exhibits suboptimal performance in environments with a different $\bar{F_c}$. For example, in the case of a budget of 15, CAtNIPP trained on $F_c=1$ outperforms CAtNIPP trained on $F_c=5$ and $F_c=10$ in an environment with $F_c=1$, whereas it performs poorly in environments with $F_c=5$ and $F_c=10$. 

\vspace{-0.5ex}
$\bullet$ \textbf{Domain randomization alone is not sufficient}: CAtNIPP+DR performs intermediate to the CAtNIPP policies trained on $F_c=1, 5, 10$. This implies that \texttt{DyPNIPP} without environment dynamic prediction converges to the optimal policy for averaged environment dynamics, which is suboptimal for each individual environment dynamic.

\vspace{-0.5ex}
$\bullet$ \textbf{\texttt{DyPNIPP} is robust against the variation in environment dynamics}: \texttt{DyPNIPP} outperforms the baselines on both metrics. In addition, \texttt{DyPNIPP} shows similar covariance trace performance regardless of the underlying fuel distribution. These are strong pieces of evidence supporting the \textbf{robustness} of our approach against variations in environment dynamics.


\subsubsection{Performance Comparison of Variation in  number of fires} We additionally study the effect on our policy when tested in environments with different number of fires, each randomly originating at different timestamps within an episode. We evaluate \texttt{DyPNIPP} and CAtNIPP + DR, both trained in an environment with up to 3 fires, on environments with 1, 3, and 5 fires each, in terms of RMSE. As seen in \tableautorefname~\ref{tab:num_fire}, the proposed method outperforms CAtNIPP + DR in all considered settings, implying that \texttt{DyPNIPP} handles variations in the number of fires as well as variations in the fuel coefficient. 

\begin{figure}
  \centering
  \includegraphics[width=0.75\linewidth]{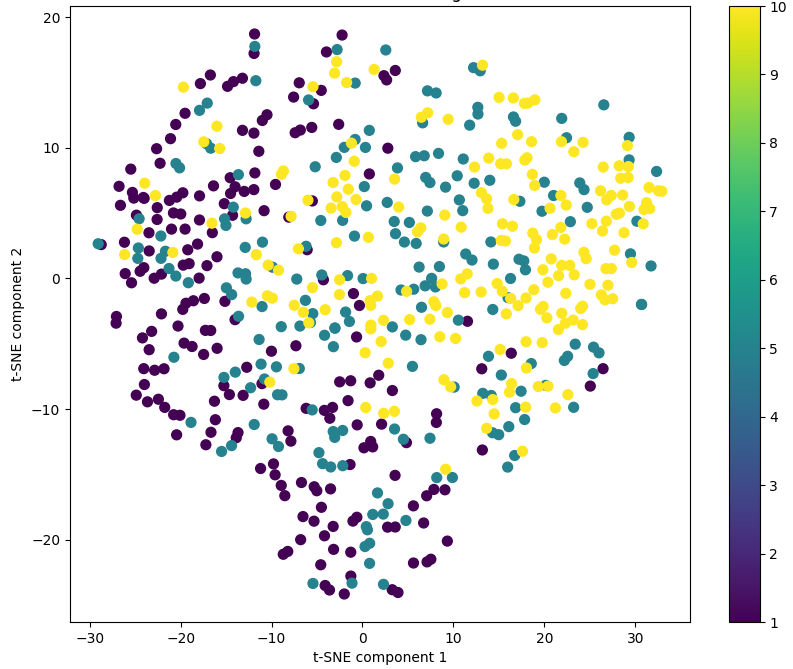}
  \caption{t-SNE plot of environment-context feature. Fuel/vegetation coefficient: 1 (purple), 5 (green), 10 (yellow)}
  \label{fig:tsne}
\end{figure}

\subsection{Analysis: Dynamics Prediction Model}

\label{sec:design_dpm}
We proposed to design the DPM to predict the belief of the next observation to capture the environment dynamics. 
To verify this, we conduct an ablation study comparing two modified versions of \texttt{DyPNIPP} that predict (1) the belief of the current observation and (2) the belief difference between the current observation and the next observation. The corresponding result is shown in \tableautorefname~\ref{tab:pred_ab}. We can see that predicting the belief of the current observation performs worse than \texttt{DyPNIPP} and the version of \texttt{DyPNIPP} that predicts the belief difference, both of which involve predicting the next observation. This is because predicting the belief of the current observation is not capable of capturing the environment dynamics. We observe that the version of \texttt{DyPNIPP} predicting the belief difference also performs well, even though its performance is slightly lower than the original \texttt{DyPNIPP}. 

Additionally, to verify that the DPM implicitly learns the environment dynamics, we investigate the latent embeddings generated by the LSTM of our belief prediction network using t-SNE~\cite{tsne}. We collect the final 16-dimensional embeddings at the end of the episode for three fuel/vegetation coefficients, $F_c \in \{1, 5, 10\}$, with 200 seeds each. These embeddings are visualized in a 2-dimensional space using t-SNE. As shown in Fig. \ref{fig:tsne}, we observe a clear progression of the latent embeddings as the fuel/vegetation coefficient increases from 1 (purple) to 5 (green), and then to 10 (yellow).

\subsection{Experimental Validation on Real Robot}
\begin{figure}
  \centering
  \includegraphics[width=0.95\linewidth]{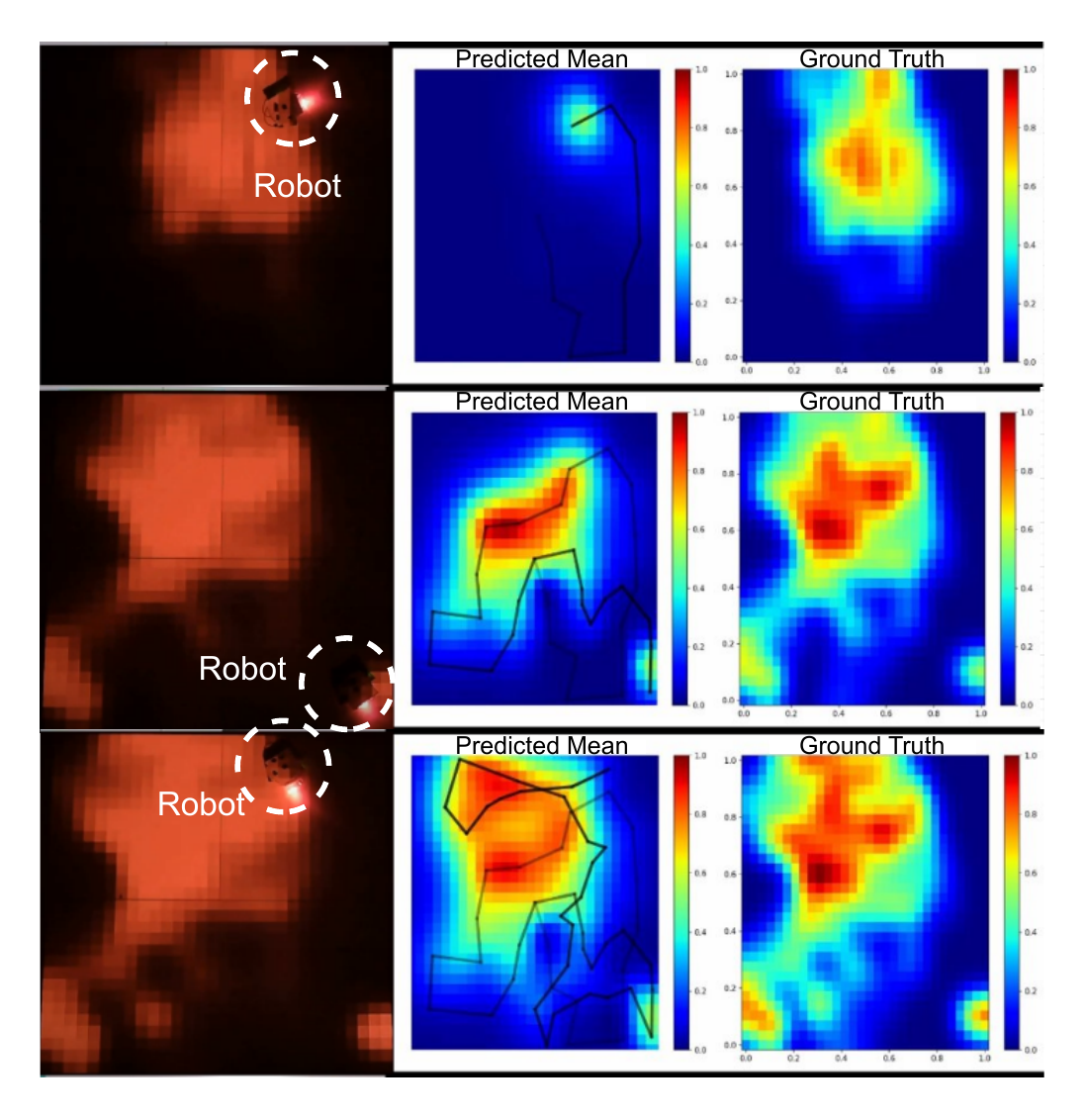}
  \caption{Experimental validation of \textbf{DyPNIPP} on the Khepara-IV robot. The left figures show the robot performing informative path planning in the arena by observing environmental phenomena at its location, with red areas indicating higher values. The right figures display the predicted environmental phenomena and ground truth, with time progressing from the first to the third row. The predicted mean improves as the robot explores more areas.}

  \label{fig:real_world}
\end{figure}

We provide experimental validation on a real robot to verify that our IPP model, trained in a simulator, can be applied in the real world. For this, we project the spatiotemporal wildfire environment onto a physical $1.5\ m \times 1.5\ m$ arena, maintaining consistency in configuration between simulation and physical deployment. We used the Khepera-IV robot, equipped with a Raspberry Pi 3 and a camera module. In this experiment, we first trained a policy in simulation and then tested it using the Khepera-IV robot and the arena. Here, the robot can only observe the intensity of the fire, which is an environmental phenomenon, at its current location. It updates its belief via GP regression using the observation, and the updated belief is used as input for the next forward pass of our policy. We use a budget of 12 m for this experiment, which corresponds to 8 units in a unit grid. It is observed that the robot trained in the simulator successfully finds a path that maximizes the information gain, as shown in Fig. \figureautorefname~\ref{fig:real_world}. Through this experiment, we also demonstrate the fast decision-making time of our policy (less than 0.25 s on an Intel Xeon CPU). This experiment shows that \texttt{DyPNIPP} can be deployed on a robot in the spatiotemporal environment. 
The full video can be accessed \href{https://drive.google.com/file/d/1_JNfKxYq_69dDNLw7P5QYMjIwOCdv561/view}{here}.

\section{CONCLUSION}

In this paper, we propose \texttt{DyPNIPP}, an RL-based framework for IPP that not only works in spatio-temporal environments but is also robust against variations in environmental dynamics. \texttt{DyPNIPP} introduces two components on top of an RL-based IPP algorithm. First, \texttt{DyPNIPP} includes domain randomization during training to encourage the agent to encounter variations in environment dynamics. Second, \texttt{DyPNIPP} includes a dynamics prediction model that predicts the belief of the next observation, allowing the agent to figure out the current environmental dynamics. We show that \texttt{DyPNIPP} is superior to existing IPP methods in terms of solution quality across diverse environmental dynamics, demonstrating its greater robustness. In addition, we provide a physical robot experiment to showcase real-time planning, highlighting its potential for real-life robotic applications.


\textbf{Limitations}~~Our subroutine algorithm's reliance on a predefined graph for traversal presents two limitations: (1) uniform sampling may leave some areas of interest unreachable due to insufficient coverage, and (2) the graph cannot adapt to unknown obstacles. We expect \texttt{DyPNIPP} to work with a dynamically sampled graph to enhance obstacle avoidance \cite{marija_3d}. Additionally, the performance of the policy relies heavily on the robot's belief, which is influenced by GP hyperparameters. Different environmental dynamics may have distinct, optimal hyperparameters, but this mapping is unknown. Future work will explore integrating GP hyperparameter optimization into the existing learning framework.





\section*{ACKNOWLEDGMENT}

This work has been supported by NSF and USDA-NIFA under AI Institute for Resilient Agriculture, Award No. 2021-67021-35329 and Department of Agriculture Award Number  2023-67021-39073.

\bibliographystyle{IEEEtran}
\bibliography{example}

\end{document}